\documentclass[10pt,twocolumn,letterpaper]{article}

\usepackage{cvpr}
\usepackage{times}
\usepackage{epsfig}
\usepackage{graphicx}
\usepackage{amsmath}
\usepackage{amssymb}

\usepackage{enumitem}
\usepackage{mmstyle}
\usepackage{makecell}
\usepackage{subcaption}
\usepackage{dsfont}
\usepackage{multirow}
\usepackage{textcomp}

\usepackage[pagebackref=true,breaklinks=true,letterpaper=true,colorlinks,bookmarks=false]{hyperref}

\cvprfinalcopy 

\graphicspath{{figures/}}

\def\cvprPaperID{2030} 

\ifcvprfinal\fi
\begin{document}
	
\title{Learning to Cluster Faces via Confidence and Connectivity Estimation}

\author{Lei Yang\textsuperscript{1},
Dapeng Chen\textsuperscript{2},
Xiaohang Zhan\textsuperscript{1},
Rui Zhao\textsuperscript{2},
Chen Change Loy\textsuperscript{3},
Dahua Lin\textsuperscript{1} \\
\textsuperscript{1}The Chinese University of Hong Kong \\
\textsuperscript{2}SenseTime Group Limited,
\textsuperscript{3}Nanyang Technological University \\
{\tt\small \{yl016, zx017, dhlin\}@ie.cuhk.edu.hk, \{chendapeng, zhaorui\}@sensetime.com, ccloy@ntu.edu.sg}
}
	
\maketitle
\thispagestyle{empty}


\begin{abstract}

Face clustering is an essential tool for exploiting the unlabeled face data,
and has a wide range of applications including face annotation and retrieval.
Recent works show that supervised clustering can result in noticeable
performance gain.
However, they usually involve heuristic
steps and require numerous overlapped subgraphs, severely restricting their
accuracy
and efficiency.
In this paper, we propose a fully learnable clustering framework without
requiring a large number of overlapped subgraphs.
Instead, we transform the clustering problem into two sub-problems.
Specifically, two graph convolutional networks, named GCN-V and GCN-E,
are designed to estimate the confidence of vertices and
the connectivity of edges, respectively. 
With the vertex confidence and edge connectivity, we can naturally organize
more relevant vertices on the affinity graph and group them into clusters.
Experiments on two large-scale benchmarks show that our method
significantly improves clustering accuracy and thus performance of the
recognition models trained on top, yet it is an order of
magnitude more efficient than existing supervised methods.
%

\end{abstract}


\section{Introduction}
\label{sec:introduction}

\begin{figure}[t]
	\centering
	\includegraphics[width=0.65\linewidth]{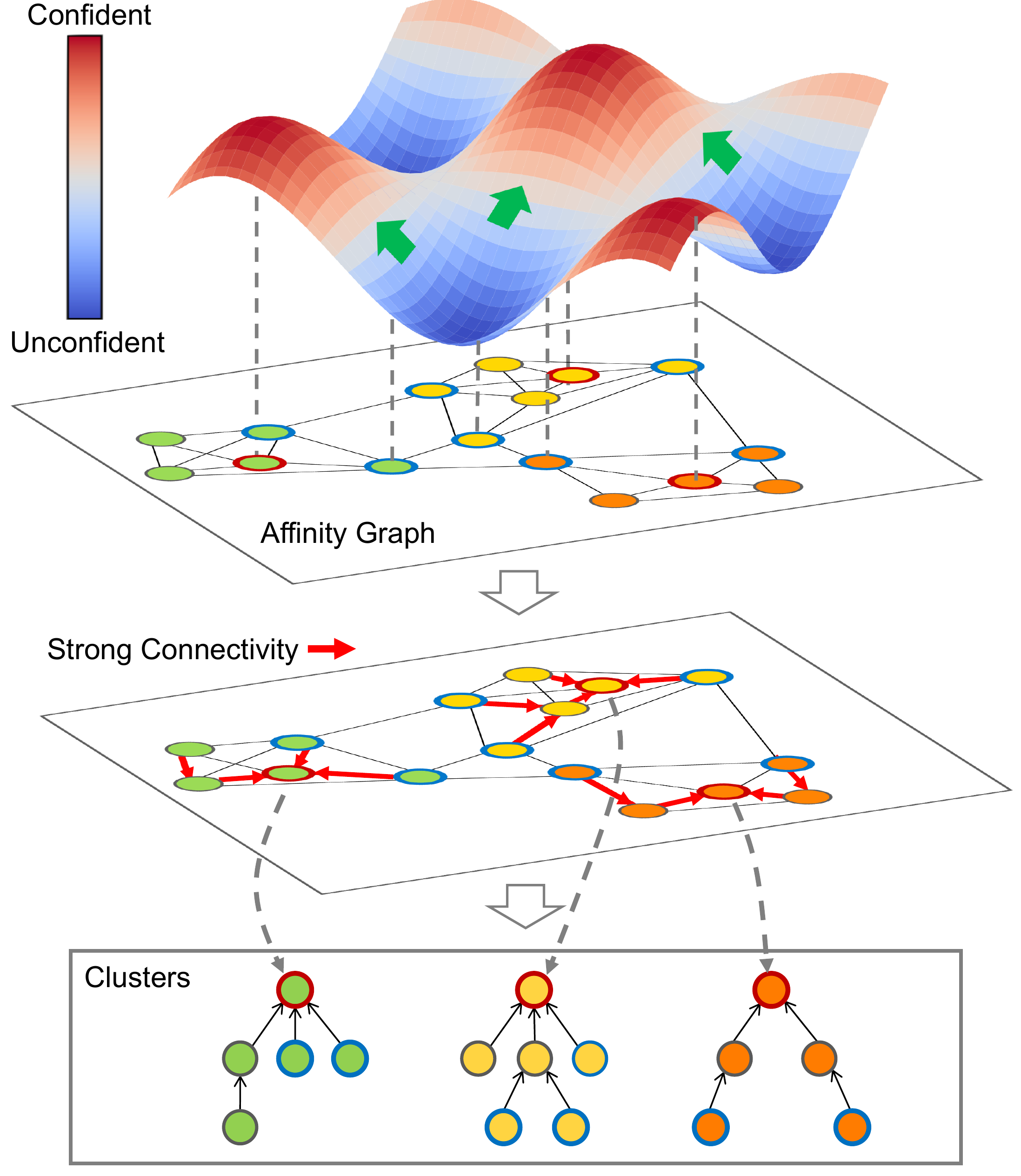}
	\caption{\small
        The core idea of our approach.
        Vertices with different colors represent different classes.
        Previous methods group all vertices in the
        box into a cluster as they are densely connected,
        while our approach, learning to estimate the confidence of
        belonging to a specific class, is able to detect unconfident
        vertices that lie among multiple classes.
        With the estimated vertex confidence, we further
        learn to predict the edge connectivity.
        By connecting each vertex to a neighbor with higher
        confidence and strongest connection,
        we partition the affinity graph into trees,
		each of which naturally represents a cluster.
    }
	\label{fig:teaser}
\end{figure}


Thanks to the explosive growth of annotated face datasets
~\cite{klare2015pushing,guo2016ms,kemelmacher2016megaface},
face recognition has witnessed great progress in recent years
~\cite{sun2014deep,schroff2015facenet,wang2018cosface,
deng2018arcface,zhang2018accelerated}.
Along with this trend, the ever-increasing demand for annotated data has
resulted in prohibitive annotation costs.
To exploit massive unlabeled face images, recent studies
~\cite{he2018merge,zhan2018consensus,wang2019linkage,yang2019learning}
provide a promising clustering-based pipeline and demonstrate its
effectiveness in improving the face recognition model.
They first perform clustering to generate ``pseudo labels'' for unlabeled
images and then leverage them to train the model in a supervised way.
The key to the success of these approaches lies in an effective face clustering
algorithm.


Existing face clustering methods roughly fall into two
categories, namely, unsupervised methods and supervised methods.
Unsupervised approaches, such as  K-means~\cite{lloyd1982least} and
DBSCAN~\cite{ester1996density}, rely on specific assumptions and lack the
capability of coping with the complex cluster structures in real-world datasets.
To improve the adaptivity to different data, supervised clustering methods have
been proposed~\cite{wang2019linkage,yang2019learning} to learn the cluster
patterns. Yet, both accuracy and efficiency are far from satisfactory.
In particular, to cluster with the large-scale face data, existing supervised
approaches organize the data with numerous small subgraphs, leading to
two main issues. First, processing subgraphs involves heuristic steps based on
simple
assumptions. Both subgraph generation~\cite{yang2019learning} and prediction
aggregation~\cite{wang2019linkage} depend on heuristic procedures,
thus limiting their performance upper bound. Furthermore, the subgraphs
required by these approaches are usually highly overlapped,
incurring excessive redundant computational costs.


We therefore seek an algorithm that learns to cluster more
\emph{accurately} and \emph{efficiently}.
For higher accuracy, we desire to make all components
of the framework learnable, moving beyond the limitations of heuristic
procedures. On the other hand, to reduce the redundant computations,
we intend to reduce the number of required subgraphs.
Previous works~\cite{zhan2018consensus,wang2019linkage,yang2019learning} have shown that
clusters on an affinity graph usually have some structural patterns.
We observe that such structural patterns are mainly originated from two sources, namely
\emph{vertices} and \emph{edges}.
Intuitively,
connecting each vertex to a neighbor, which has higher confidence of
belonging to a specific class, can deduce a number of trees from the affinity
graph.
The obtained trees naturally form connected components as clusters.
Based on this motivation, we design a fully learnable clustering approach,
without requiring numerous subgraphs, thus
leading to high accuracy and efficiency.

%
%
Particularly, we transform the clustering problem into two sub-problems.
One is to estimate the confidence of a vertex, which measures
the probability of a vertex belonging to a specific class.
The other is
to estimate the edge connectivity, which indicates the probability of two
vertices belonging to the same class. 
With the vertex confidence and edge
connectivity, we perform clustering in a natural way, \ie,
each vertex is
connected to a vertex with higher confidence and strongest connectivity.
As Figure~\ref{fig:teaser} illustrates, each vertex finds an edge connected
to a vertex with higher confidence, and the vertices that finally connected to
the same vertex belong to the same cluster.

Two learnable components, namely,  a \emph{confidence estimator} and a
\emph{connectivity estimator} are proposed to estimate the vertex confidence
and
edge connectivity, respectively. Both components are based on a GCN to learn
from the data, denoted by GCN-V (for vertex confidence) and GCN-E (for edge connectivity).
Specifically, GCN-V takes the entire graph as input and simultaneously
estimates confidence for all vertices. GCN-E takes the graph constructed from a
local candidate set as input and evaluates the possibility of two vertices
belonging to the same class.


%

%
%


The experiments demonstrate that our approach not only significantly
accelerates the existing supervised methods by an order of magnitude,
but also outperforms the recent state of the art~\cite{yang2019learning} under
two F-score metrics on $5M$ unlabeled data.
The main contributions lie in three aspects:
(1) We propose a novel framework that formulates clustering as an estimation of
confidence and connectivity, both based on learnable components.
(2) Our approach is an order of magnitude faster than existing
learning-based methods.
(3) The proposed method achieves state-of-the-art performance on both large-scale
face clustering and fashion clustering.
The discovered clusters boost the face recognition model to a level
that is comparable to its
supervised
counterparts.

\section{Related Work}
\label{sec:related}

\begin{figure*}[t]
	\centering
	\includegraphics[width=0.8\linewidth]{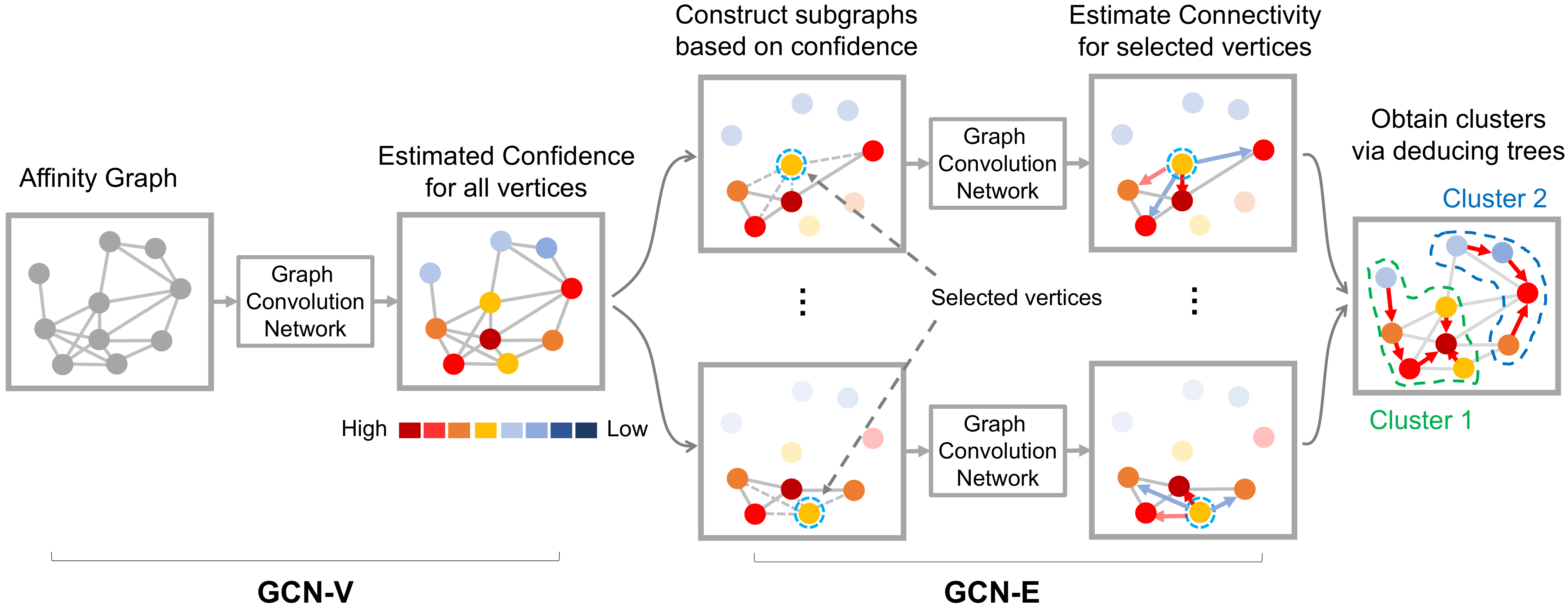}
	\caption{\small
        Overview of the proposed clustering framework.
    }
	\label{fig:pipeline}
\end{figure*}

\paragraph{Unsupervised Face Clustering.}
%
%
With the emergence of deep learning, recent works primarily adopt deep features
from a CNN-based model and focus on the design of similarity metrics.
%
%
%
Otto~\etal~\cite{otto2018clustering} proposed an approximate rank-order metric.
Lin~\etal~\cite{lin2018deep} introduced minimal covering spheres of
neighborhoods as the similarity metric.
%
Besides methods designed specifically for face clustering, classical
clustering algorithms can also be applied to face clustering.
%
Density-based clustering is the most related approach.
DBSCAN~\cite{ester1996density} computed empirical density and designated
clusters as dense regions in the data space.
%
%
OPTICS~\cite{ankerst1999optics} adopted similar concepts
and addresses the ordering of data points.

The proposed method shares common intuitions with the
density-based clustering, \ie, computing the ``density'' for each
sample~\cite{ester1996density} and focusing on the relative order
between samples~\cite{ankerst1999optics}.
Yet, our method differs substantially with all the unsupervised methods above:
all components in our framework are \emph{learnable}.
This allows us to learn to capture the intrinsic structures in face clusters.

\vspace{-11pt}
\paragraph{Supervised Face Clustering.}
%
%
Recent works have shown that the introduced supervised information in face
clustering leads to considerable performance gains.
Zhan~\etal~\cite{zhan2018consensus} trained a MLP classifier to aggregate
information and thus discover more robust linkages.
Wang~\etal~\cite{wang2019linkage} further improved the linkage prediction by
leveraging GCN to capture graph context.
Both methods obtained clusters by finding connected components with
dynamic threshold.
Yang~\etal~\cite{yang2019learning} devised a partitioning algorithm to
generate multi-scale subgraphs and proposed a two-stage supervised
framework to pinpoint desired clusters therefrom.

Whereas the proposed method adopts the idea of supervised clustering, it
differs from two key aspects:
(1) Unlike previous supervised
methods~\cite{zhan2018consensus,wang2019linkage,yang2019learning},
it does not rely on heuristic algorithms for pre-processing or post-processing.
Instead, all components of the proposed framework are learnable and can
potentially achieve higher accuracy.
(2) It is more efficient in design.
Existing methods rely on a large number of subgraphs for pinpointing
clusters.
\cite{wang2019linkage} predicted all connections around each vertex, where two
nearby vertices are likely to have highly overlapped neighborhoods, and thus
there are redundant computational costs.
\cite{yang2019learning} produced multi-scale
subgraphs for detection and segmentation, the number of
which is usually several times larger than the number of clusters.
In contrast, the proposed method adopts an efficient subgraph-free strategy to
estimate the vertex confidence and concentrates on a small portion of
neighborhoods for connectivity prediction.

\vspace{-11pt}
\paragraph{Graph Convolutional Networks.}
Graph Convolutional Networks (GCNs)~\cite{kipf2017semi} have been successfully
applied to various tasks~\cite{kipf2017semi,hamilton2017inductive,
van2017graph,yan2018spatial,yan2019convolutional}.
%
Some recent efforts extend GCN to handle large-scale graphs.
GraphSAGE~\cite{hamilton2017inductive} sampled a fixed number of
neighbors in each layer for aggregation.
FastGCN~\cite{chen2018fastgcn} further reduced computational cost by
sampling vertices rather than neighbors.
%
%
In this paper, we draw on the strong expressive power of graph convolutional
networks, to learn vertex confidence on the massive affinity graph and
edge connectivity on the local subgraphs.


\section{Methodology}
\label{sec:methodology}

In large-scale face clustering, supervised approaches demonstrate their
effectiveness in handling complex cluster patterns, but their accuracy is
limited by some hand-crafted components and their efficiency suffers from the
requirement of numerous highly overlapped subgraphs.
Therefore, how to cluster accurately and efficiently remains a problem.
To address the challenge, we propose an efficient alternative in which
all components are learnable.
Specifically, we formulate clustering as a procedure of estimating
vertex confidence and edge connectivity on an affinity graph, and then
partition
the graph into clusters by connecting each vertex to neighbors with
higher confidence and connectivity.


\subsection{Framework Overview}

Given a dataset,
we extract the feature for each image from a trained CNN, forming a
feature set $\cF=\{\vf_i\}_{i=1}^{N}$, where $\vf_{i} \in \mathbb{R}^D$.
$N$ is the number of images and $D$ denotes the feature dimension.
The affinity between sample $i$ and sample $j$ is denoted as $a_{i,j}$,
which is the cosine similarity between $\vf_{i}$ and $\vf_{j}$.
According to the
affinities, we represent the dataset with a $K$NN affinity graph $\cG = (\cV,
\cE)$, where each image is a vertex belonging to $\cV$ and is connected to its
$K$ nearest neighbors, forming $K$ edges belonging to $\cE$. The constructed
graph can be expressed as a vertex feature matrix $\mF \in \mathbb{R}^{N \times
D}$ and a symmetric adjacency matrix $\mA\in \mathbb{R}^{N \times N}$, where
$a_{i,j} = 0$ if $v_{i}$ and $v_{j}$ are not connected.

To perform clustering by learning the structural patterns from vertices
and edges,
we decompose the clustering into
two sub-problems. One is to predict the \emph{confidence} of the vertices.  The
confidence is to determine whether a vertex belongs to a specific
class. Intuitively, a vertex with high confidence usually lies in the place
where the vertices are densely distributed and belong to the same class, while
the vertices with low confidence are likely to be on the boundary among several
clusters. The other is sub-problem to predict the \emph{connectivity} of the edges.
The edge with high connectivity indicates the two connected samples tend to
belong to the same class.
With the vertex confidence and the edge connectivity in the
affinity graph, clustering can be performed in a simple way by finding a
directed path from vertices with lower confidence to those with higher
confidence.
This process naturally forms a number of trees isolated from each other, thus
readily partitioning the graph into clusters.
We refer to this process as \emph{tree-based partition}.

The key challenge for the proposed method remains in how to estimate vertex
confidence
and edge connectivity. As shown in Figure~\ref{fig:pipeline}, our
framework consists of two learnable modules, namely
\emph{Confidence Estimator} and \emph{Connectivity Estimator}.
The former estimates the vertex confidence based on GCN-V,
while the latter predicts the edge connectivity based on GCN-E. 
Specifically, GCN-V takes the entire affinity graph as input and
simultaneously estimates confidence for all vertices.
GCN-E takes the graph constructed from a candidate set as input
and evaluates the confidence of two vertices belonging to the same class.
According to the output of these two modules, we perform our tree-based
partition to obtain clusters.
%

\subsection{Confidence Estimator}
\label{sec:gcn_v}

Similar to anchor-free methods in
object detection~\cite{zhou2019objects,duan2019centernet}, where they use
heatmap to indicate the possibility that an object appears in the corresponding
area of an image,
the confidence estimator aims to estimate a value for each vertex, thereby
indicating whether there is a specific class on the corresponding area of an affinity graph.


As real-world datasets usually have large intra-class variations, each image
may have different confidence values even when they belong to the same class.
For an image with high confidence, its neighboring images tend to belong to the
same class while an image with low confidence is usually adjacent to the
images from the other class. Based on this observation, we can define the
confidence $c_{i}$ for each vertex based on the labeled images in the
neighborhood:
\begin{equation}
    c_i = \frac{1}{|\cN_i|} \sum_{v_j \in \cN_i} (\mathds{1}_{y_j = y_i} -
    \mathds{1}_{y_j \ne y_i}) \cdot a_{i, j},
    \label{eq:v_importance}
\end{equation}
where $\cN_i$ is the neighborhood of $v_i$,  $y_i$ is the ground-truth label
of $v_i$, and $a_{i,j}$ is the affinity between $v_{i}$ and $v_{j}$. The
confidence measures whether the neighbors are close and from the same class.
Intuitively, vertices with dense and pure connections have high confidence,
while vertices with sparse connections or residing in the boundary among
several clusters have low confidence.
We investigate some different designs of confidence in Sec.~\ref{sec:v_det}.


\vspace{-7pt}
\paragraph{Design of Confidence Estimator.}
We assume that vertices with similar confidence have similar
structural patterns.
To capture such patterns, we learn a graph convolutional
network~\cite{kipf2017semi}, named GCN-V, to estimate confidence of
vertices.
Specifically, given the adjacency matrix $\mA$ and the vertex
feature matrix $\mF$ as input, the GCN predicts confidence for each
vertex.
The GCN consists of $L$ layers and the computation of each layer can be
formulated as:
\begin{equation}
	\mF_{l+1} = \sigma\left( g(\tilde{\mA}, \mF_l)\mW_l \right),
	\label{eq:gcn}
\end{equation}
where
$\tilde{\mA} = \tilde{\mD}^{-1} (\mA + \mI)$ and
$\tilde{\mD}_{ii} = \sum_j (\mA + \mI)_j$
is a diagonal degree matrix. The feature embedding of the input layer
$\mF_0$ is set with the feature matrix $\mF$, and $\mF_l$ contains the
embeddings at $l$-th layer. $\mW_l$ is a trainable matrix to transform the
embeddings into a new space. $\sigma$ is a nonlinear activations (\emph{ReLU}
in
this work).
To leverage both input embeddings and embeddings after neighborhood
aggregation to learn the transformation matrix,
we define $g(\cdot, \cdot)$ as the concatenation of them:
\begin{equation}
	g(\tilde{\mA}, \mF_l) = [(\mF_l)^{\top} ,  (\tilde{\mA}\mF_l)^{\top}]^{\top}.
	\label{eq:op}
\end{equation}
Such definition has been proven to be more effective than
simply taking weighted average of the embedded feature of
neighbors around each vertex~\cite{wang2019linkage}.
Based on the output embedding of the $L$-th layer,
\ie, $\mF_L$, we employ a fully-connected layer to predict the
confidence of the vertices.
\begin{equation}
  \vc' = \mF_L \mW + \vb,
\end{equation}
where $\mW$ is trainable regressor and $\vb$ is trainable bias.
The predicted confidence of $v_{i}$ can be taken from the corresponding element
in $\mathbf{c}'$, denoted by $c'_{i}$.

\vspace{-7pt}
\paragraph{Training and Inference.}
Given a training set with class labels, we can obtain the ground-truth
confidence following Eq.~\ref{eq:v_importance} for each vertex.
Then we train GCN-V, with the objective to minimize the
\emph{mean square error(MSE)} between ground truth and predicted
scores, which is defined by:
\begin{equation}
\cL_{V} = \frac{1}{N}\sum_{i=1}^{N} |c_{i} - c_{i}'|^{2}
\label{eq:v_loss}
\end{equation}
During inference, we use the trained GCN-V to predict the
confidence of each vertex.
The obtained confidence is used in two ways.
First, they are used in the next module
to determine whether the connectivity of an edge needs to be predicted, thus
significantly reduces the computational cost.
Furthermore, they are used in the final clustering to provide partial orders between vertices.

\vspace{-7pt}
\paragraph{Complexity Analysis.}
The main computational cost lies in the graph convolution
(Eq.~\ref{eq:gcn}).
Since the built graph is a $K$NN graph with $K \ll N$, the affinity matrix $\mA$
is a highly sparse matrix.
Therefore, the graph convolution can be efficiently implemented as the
sparse-dense
matrix multiplication, yielding a complexity
$\cO(|\cE|)$~\cite{kipf2017semi}.
As the number of edges $|\cE|$ of the sparse matrix is bounded by $NK$,
the inference complexity is linear in the number of vertices as $K \ll N$.
This operation can be scaled to a very large setting by sampling
neighbors or sampling vertices~\cite{hamilton2017inductive,chen2018fastgcn}.
Empirically, a 1-layer GCN takes $37G$ CPU Ram and $92s$ with $16$ CPU
on a graph with $5.2M$ vertices for inference.
%
%

\subsection{Connectivity Estimator}
For a vertex $v_i$,
neighbors with confidence larger than $c_i$ indicate they
are more confident to belong to a specific class.
To assign $v_i$ to a specific class,
an intuitive idea is to connect $v_i$ to neighbors from the same class
with larger confidence.
However, neighbors with larger confidence do not necessarily belong to the same class.
We therefore introduce the connectivity estimator, named GCN-E, to
measure the pairwise relationship based on the local graph structures.

\vspace{-7pt}
\paragraph{Candidate set.}
Given the predicted vertex confidence, we first construct a candidate
set $\cS$ for each vertex.
\begin{equation}
    \cS_i = \{ v_{j} | c'_j > c'_i, v_j \in \cN_i \}.
    \label{eq:candidate_set}
\end{equation}
The idea of candidate set is to select edges connected to neighbors more
confident to belong to a cluster, and $\cS_{i}$ only contains the vertices with
higher confidence than the confidence of $v_{i}$.


\vspace{-7pt}
\paragraph{Design of Connectivity Estimator.}
GCN-E shares similar GCN structures with GCN-V.
The main difference lies in:
(1) Instead of operating on the entire graph $\cG$, the input of
GCN-E is a subgraph $\cG(\cS_i)$ containing all vertices in $\cS_{i}$;
(2) GCN-E outputs a value for each vertex on $\cG(\cS_{i})$ to indicate how
likely it shares the same class with $v_i$.

More specifically, the subgraph $\cG(\cC_i)$ can be represented by the affinity
matrix $\mA(\cS_{i})$ and the vertex feature matrix $\mF(\cS_{i})$.
We subtract $\vf_i$ from each row of the feature matrix $\mF(\cS_{i})$ to encode the relationship
between $\cS_i$ and $v_i$, and the obtained feature matrix is
denoted by $\bar{\mF}(\cS_{i})$. The transformation in GCN-E can be therefore
represented by:
\begin{equation}
\bar{\mF}_{l+1} = \sigma\left( g(\tilde{\mA}(\cC_{i}),
\bar{\mF}_l(\cC_{i}))\mW'_l \right),
\end{equation}
where $\sigma$, $g(\cdot)$ and $\tilde{\mA}(\cS_{i})$ are defined similar to
those in Eq.~\ref{eq:gcn}. $\mW'_l$ is the parameter of GCN-E in the $l$-th
layer. Based on the
output embedding of the $L$-th layer, we obtain the connectivity for
each vertex in $\cS_{i}$ by a fully-connected layer. As the connectivity
reflects the relationship between two vertices, we use $r'_{i,j}$ to
indicate the predicted connectivity between $v_{i}$ and $v_{j}$.

\vspace{-7pt}
\paragraph{Training and Inference.}
Given a training set with class labels,
for a vertex $v_i$, if a neighbor $v_j$ shares the same label with the $v_i$,
the connectivity is set to $1$, otherwise it is $0$.
\begin{equation}
	r_{i, j} =
	\begin{cases}
    1, & y_i = y_j \\
    0, & y_i \ne y_j
	\end{cases},
	v_j \in \cC_i,
    \label{eq:e_importance}
\end{equation}
We aim to predict the connectivity that reflects whether two vertices belong to
the same class.
Similar to Eq.~\ref{eq:v_loss} in GCN-V, we also use vertex-wise MSE
loss to train GCN-E.
\begin{equation}
\cL_{E}(\cC_i) = \sum_{v_j\in \cC_{i}}|r_{i,j} - r'_{i,j}|^{2}
\end{equation}
To accelerate the training and inference procedures, we only apply GCN-E to
a small portion of vertices with large estimated confidence, as they
potentially
influence more successors than vertices with small confidence do.
We denote the portion of vertices using GCN-E as $\rho$.
For other vertices, they simply connect to their $M$ nearest neighbors
in the candidate set, indicating they connect to neighbors with
top-$M$ largest similarities and higher confidence.
$M=1$ leads to the tree-based partition strategy, while $M>1$ produces directed
acyclic graphs as clusters.
%
%
Empirical results indicate that $M=1, \rho=10\%$ can already bring considerable
performance gain (see Sec.~\ref{sec:e_det}).
%

\vspace{-7pt}
\paragraph{Complexity Analysis.}
The idea of connectivity estimator shares similar spirits to
~\cite{wang2019linkage},
where they evaluated how likely each vertex on a subgraph connects to the
center vertex.
Although the complexity of ~\cite{wang2019linkage} is linear with $N$, applying
a GCN on the neighborhood of each vertex incurs excessive
computational demands.
The proposed GCN-E has two key designs to be much more efficient:
(1) We only predict linkages in the candidate set, an effort that potentially
involves fewer neighbors for each vertex and does not need to manually select
the number of hops and the number of neighbors for each hop.
(2) With the estimated vertex confidence, we are able to focus on
a small portion of vertices with high confidence.
With these two important designs, we achieve a speedup
over ~\cite{wang2019linkage} by an order of magnitude.


\section{Experiments}
\label{sec:experiment}

\begin{table*}[]
\centering
\caption{
Comparison on face clustering with different numbers of unlabeled images.
(MS-Celeb-1M)}
\begin{tabular}{l|cc|cc|cc|cc|cc|c}
\hline
\#unlabeled & \multicolumn{2}{c|}{584K} & \multicolumn{2}{c|}{1.74M} &
\multicolumn{2}{c|}{2.89M} & \multicolumn{2}{c|}{4.05M} &
\multicolumn{2}{c|}{5.21M} & \multirow{2}{*}{Time}\\ \cline{1-11}
Method $/$ Metrics & $F_P$ & $F_B$ & $F_P$ & $F_B$ & $F_P$ & $F_B$ 
	& $F_P$ & $F_B$ & $F_P$ & $F_B$ & \\\hline
K-means~\cite{lloyd1982least,sculley2010web} & 79.21 & 81.23 & 73.04 &
75.2 & 69.83 & 72.34 & 67.9 & 70.57 & 66.47 & 69.42 & 11.5h \\
HAC~\cite{sibson1973slink} & 70.63 & 70.46 & 54.4 & 69.53 & 11.08 & 68.62 &
1.4 & 67.69 & 0.37 & 66.96 & 12.7h\\
DBSCAN~\cite{ester1996density} & 67.93 & 67.17 & 63.41 & 66.53
	& 52.5 & 66.26 & 45.24 & 44.87 & 44.94 & 44.74 & \textbf{1.9m} \\
ARO~\cite{otto2018clustering} & 13.6 & 17 & 8.78 & 12.42 & 7.3 & 10.96
& 6.86 & 10.5 & 6.35 & 10.01 & 27.5m \\
CDP~\cite{zhan2018consensus} & 75.02 & 78.7 & 70.75 & 75.82 & 69.51
	& 74.58 & 68.62 & 73.62 & 68.06 & 72.92 & 2.3m \\
L-GCN~\cite{wang2019linkage} & 78.68 & 84.37 & 75.83 & 81.61 & 74.29 &
80.11 & 73.7 & 79.33 & 72.99 & 78.6 & 86.8m \\
LTC~\cite{yang2019learning} & 85.66 & 85.52 & 82.41 & \textbf{83.01} & 80.32 &
81.1 & 78.98 & 79.84 & 77.87 & 78.86 & 62.2m \\
\hline\hline
\textbf{Ours (V)} & 87.14 & 85.82 & 83.49 & 82.63 & 81.51 & 81.05 &
79.97 & 79.92 & 78.77 & 79.09 & 4.5m
\\
\textbf{Ours (V + E)} & \textbf{87.93} & \textbf{86.09} &
\textbf{84.04} & 82.84 & \textbf{82.1} & \textbf{81.24} &
\textbf{80.45} & \textbf{80.09} & \textbf{79.3} & \textbf{79.25} & 11.5m \\ \hline
\end{tabular}
\label{tab:exp_ms1m}
\end{table*}

\begin{table}[]
\centering
\caption{Performance on DeepFashion clustering.}
\begin{tabular}{c|c|cc|c}
\hline
Methods & \#clusters & $F_P$ & $F_B$ & Time \\\hline\hline
K-means~\cite{lloyd1982least} & 3991 & 32.86 & 53.77 & 573s \\
HAC~\cite{sibson1973slink} & 17410 & 22.54 & 48.77 & 112s \\
DBSCAN~\cite{ester1996density} & 14350 & 25.07 & 53.23 & 2.2s \\
MeanShift~\cite{cheng1995mean} & 8435 & 31.61 & 56.73 & 2.2h\\
Spectral~\cite{ho2003clustering} & 2504 & 29.02 & 46.4 & 2.1h \\
ARO~\cite{otto2018clustering} & 10504 & 26.03 & 53.01 & 6.7s  \\
CDP~\cite{zhan2018consensus} & 6622 & 28.28 & 57.83 & \textbf{1.3s} \\
L-GCN~\cite{wang2019linkage} & 10137 & 28.85 & 58.91 & 23.3s \\
LTC~\cite{yang2019learning} & 9246 & 29.14 & 59.11 & 13.1s  \\
\hline\hline
\textbf{Ours (V)} & 4998 & 33.07 & 57.26 & 2.5s \\
\textbf{Ours (V + E)} & 6079 & \textbf{38.47} & \textbf{60.06} & 18.5s \\\hline
\end{tabular}
\label{tab:exp_deepfashion}
\end{table}

\subsection{Experimental Settings}
\label{sec:exp_setting}

\noindent\textbf{Face clustering.}
MS-Celeb-1M~\cite{guo2016ms} is a large-scale face recognition dataset
consisting of $100K$ identities, and each identity has about $100$
facial images.
We adopt the widely used annotations from ArcFace~\cite{deng2018arcface},
yielding a reliable subset that contains $5.8M$ images from $86K$ classes.
We randomly split the cleaned dataset into $10$ parts with an almost
equal number of identities. Each part contains $8.6K$ identities
with around $580K$ images.
We randomly select $1$ part as labeled data and the other $9$ parts
as unlabeled data.

\noindent\textbf{Fashion clustering.}
We also evaluate the effectiveness of our approach for datasets beyond
the face images.
We test on a large subset of DeepFashion~\cite{liuLQWTcvpr16DeepFashion},
namely In-shop Clothes Retrieval, which is very \emph{long-tail}.
Particularly, we mix the training features and testing features in the original
split, and randomly sample $25,752$ images from $3,997$ categories for
training and the other $26,960$ images with $3,984$ categories for testing.
Note that fashion clustering is also regarded as an open set problem and
there is no overlap between training categories and testing categories.

\noindent\textbf{Face recognition.}
We evaluate face recognition model on MegaFace~\cite{kemelmacher2016megaface},
which is the largest benchmark for face recognition.
It includes a probe set from FaceScrub~\cite{ng2014data} with $3,530$ images
and a gallery set containing $1M$ images.

\noindent\textbf{Metrics.} We assess the performance on both \emph{clustering}
and \emph{face recognition}.
Face clustering is commonly evaluated by two
metrics~\cite{shi2018face,wang2019linkage,yang2019learning}, namely
\emph{Pairwise F-score} and \emph{BCubed
F-score}~\cite{amigo2009comparison}.
The former emphasizes on large clusters as the number of pairs grows
quadratically with cluster size, while the latter weights clusters
according to their cluster size.
Both metrics are the harmonic mean of precision and recall,
referred to as $F_P$ and $F_B$, respectively.
Face recognition is evaluated with \emph{face identification}
benchmark in MegaFace.
We adopt top-$1$ identification hit rate in MegaFace, which is to
rank the top-$1$ image from the $1M$ gallery images and compute the
top-$1$ hit rate.

\noindent\textbf{Implementation Details.}
To construct the $K$NN affinity graph, we set $K=80$ for MS1M and $K=5$ for
DeepFashion.
Since GCN-V operates on a graph with millions of vertices, we
only use $1$-layer GCN to reduce the computational cost.
For GCN-E, it operates on a neighborhood with no more than $K$ vertices,
and thus we use $4$-layer GCN to increase its expressive power.
For both datasets, momentum SGD is used with a start learning rate
$0.1$ and weight decay $1e^{-5}$.
To avoid the situation where there is no correct neighbor for
connection, we set a threshold $\tau$ to cut off the edges with small
similarities. $\tau$ is set to $0.8$ for all settings.

\subsection{Method Comparison}

\subsubsection{Face Clustering}

We compare the proposed method with a series of clustering baselines.
These methods are briefly described below.

\noindent\textbf{(1) K-means~\cite{lloyd1982least},}
the commonly used clustering algorithm.
%
For $N \ge 1.74M$, we use mini-batch K-means,
yielding a comparable result but significantly shorted running time.

%
\noindent\textbf{(2) HAC~\cite{sibson1973slink},} the method
hierarchically merges close clusters based on some criteria in a bottom-up
manner.

%
\noindent\textbf{(3) DBSCAN~\cite{ester1996density}}
extracts clusters based on a designed density criterion and
leaves the sparse background as noises.

\noindent\textbf{(4) MeanShift~\cite{comaniciu1999mean}} pinpoints clusters
which contain a set of points converging to the same local optimal.

\noindent\textbf{(5) Spectral~\cite{ng2002spectral}}
partitions the data into connected components based on spectrum of the similarity matrix.

\noindent\textbf{(6) ARO~\cite{otto2018clustering}}
performs clustering with an approximate nearest neighbor search and a
modified distance measure.

\noindent\textbf{(7) CDP~\cite{zhan2018consensus},} a graph-based clustering
algorithm, which exploits more robust pairwise relationship. 

\noindent\textbf{(8) L-GCN~\cite{wang2019linkage},}
a recent supervised method that adopts GCNs to exploit graph context for
pairwise prediction.

\noindent\textbf{(9) LTC~\cite{yang2019learning},}
another recent supervised method that formulates clustering as a detection and
segmentation pipeline.

\noindent\textbf{(10) Ours (V),} the proposed method that
applies GCN-V on the entire graph and obtains clusters through connecting
each vertex to its nearest neighbor in the candidate set.

\noindent\textbf{(11) Ours (V + E),} the proposed method that 
employs GCN-E on top of GCN-V to estimate the connectivity
and obtain clusters by connecting each vertex to the most connective
neighbors in the candidate set.

\vspace{-7pt}
\paragraph{Results}
For all methods, we tune the corresponding hyper-parameters and report the best
results.
%
The results in Table~\ref{tab:exp_ms1m} and Table~\ref{tab:exp_deepfashion} show:
(1) Given the ground-truth number of clusters, K-means achieves a high F-score.
However, the performance is influenced greatly by the number of
clusters, making it hard to employ when the number of clusters is unknown.
(2) HAC does not require the number of clusters but the iterative merging
process involves a large computational budget.
Even using a fast implementation~\cite{mullner2013fastcluster},
it takes nearly 900 hours to yield results when $N$ is $5.21M$.
(3) Although DBSCAN is very efficient, it assumes that density
among different clusters is similar, which may be the reason for severe
performance drop when scaling to large settings.
(4) MeanShift yields a good result on fashion clustering but takes
a long time to converge.
%
(5) Spectral clustering also performs well but solving eigenvalue
decomposition incurs large computation and memory demand, thus limiting its application.
(6) The performance of ARO depends on the number of neighbors. With a
reasonable time budget, the performance is inferior to other methods in MS1M.
(7) CDP is very efficient and achieves a high F-score on different
datasets with different scales.
For a fair comparison, we compare with the single model version of CDP.
(8) L-GCN surpasses CDP consistently but it is an order of magnitude slower than CDP.
(9) As a recent approach to cluster face in a supervised manner,
LTC shows its advantage in large-scale clustering.
However, relying on the iterative proposal strategy, the performance gain
is accompanied by a large computational cost.
(10) The proposed GCN-V outperforms previous methods consistently.
Although the training set of GCN-V only contains $580K$ images, it generalizes
well to $5.21M$ unlabeled data, demonstrating its effectivenesss in
capturing important characteristics of vertices.
Besides, as GCN-V simultaneously predicts the confidence for all vertices,
it is an order of magnitude faster than previous supervised approaches.
(11) We apply GCN-E to $20\%$ vertices with top estimated confidence.
It brings further performance gain, especially when applied to DeepFashion.
This challenging dataset contains noisy neighborhoods, and thus it is required to
select connectivity more carefully.

\begin{figure}[t]
	\centering
	\includegraphics[width=0.8\linewidth]{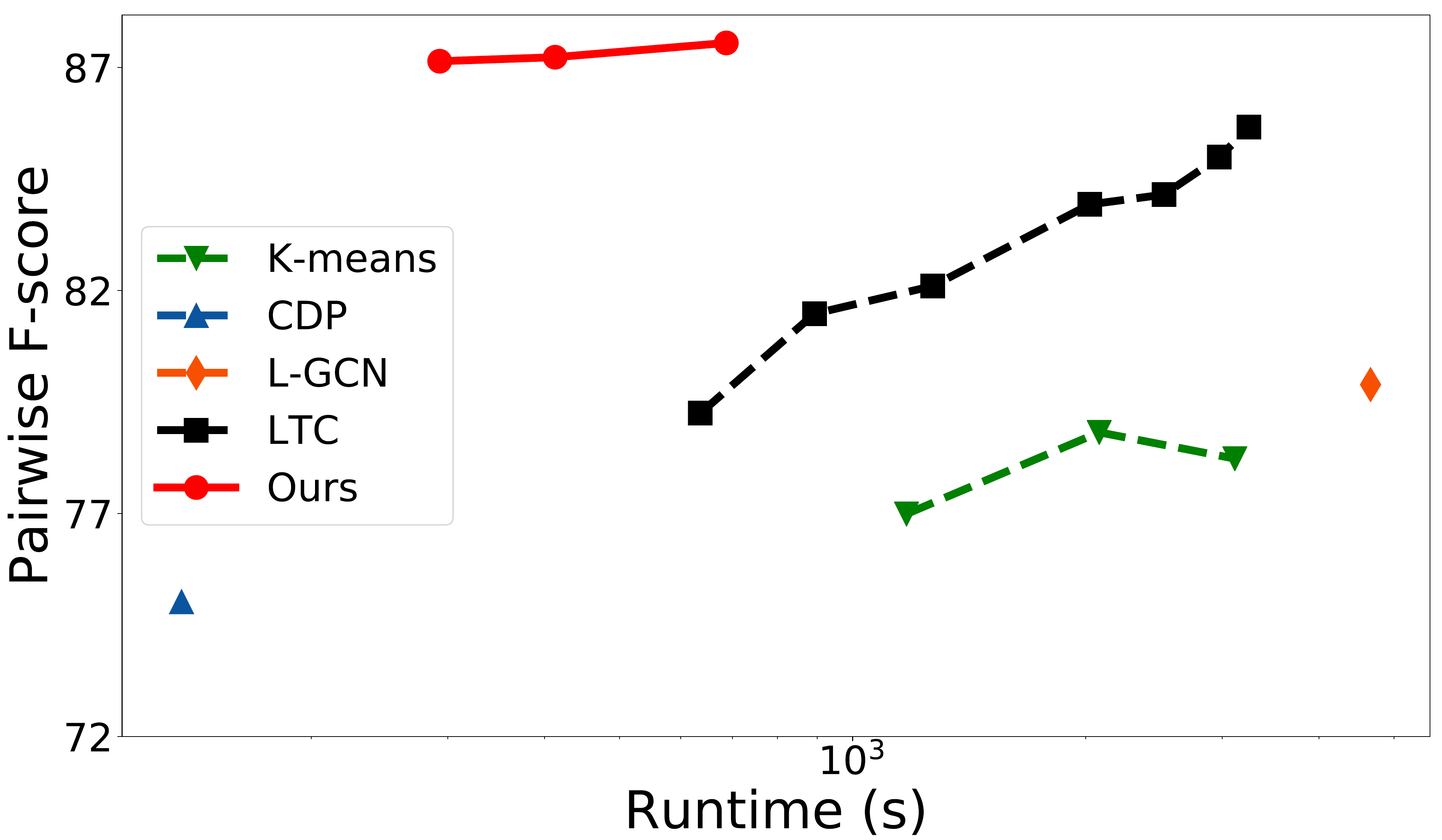}
	\caption{\small
        Pairwise F-score \vs. the runtime of different methods.
        Note that x-axis is in \emph{log-scale}.
    }
	\label{fig:exp_runtime}
\end{figure}

\vspace{-7pt}
\paragraph{Runtime Analysis}
We measure the runtime of different methods with ES-2640 v3 CPU and TitanXP.
For MS-Celeb-1M, we measure the runtime when $N=584K$.
All the compared approaches, except K-means and HAC, rely on the $K$NN graph.
To focus on the runtime of algorithms themselves, we use $1$ GPU with $16$ CPU
to
accelerate the search of $K$NN~\cite{JDH17}, which reduces the time of finding
$80$ nearest neighbors from 34min to 101s.
%
%
For all the supervised methods, we analyze their inference time.
As shown in Table~\ref{tab:exp_ms1m}, the proposed GCN-V is faster than L-GCN
and LTC by an order of magnitude.
GCN-E takes more time to predict the connectivity in the candidate sets,
but it is still several times more efficient than L-GCN and LTC.
Figure~\ref{fig:exp_runtime} better illustrates the trade-off
between accuracy and efficiency.
For LTC and Mini-batch K-means, we control the number of proposals and batch
size respectively, to yield different runtime and accuracy.
In real practice, we can leverage the idea of super vertex in LTC to further
accelerate GCN-V, and parallelize GCN-E to estimate
connectivity for different vertices simultaneously.

\vspace{-13pt}
\subsubsection{Face Recognition}

\begin{figure}[t]
	\centering
	\includegraphics[width=0.8\linewidth]{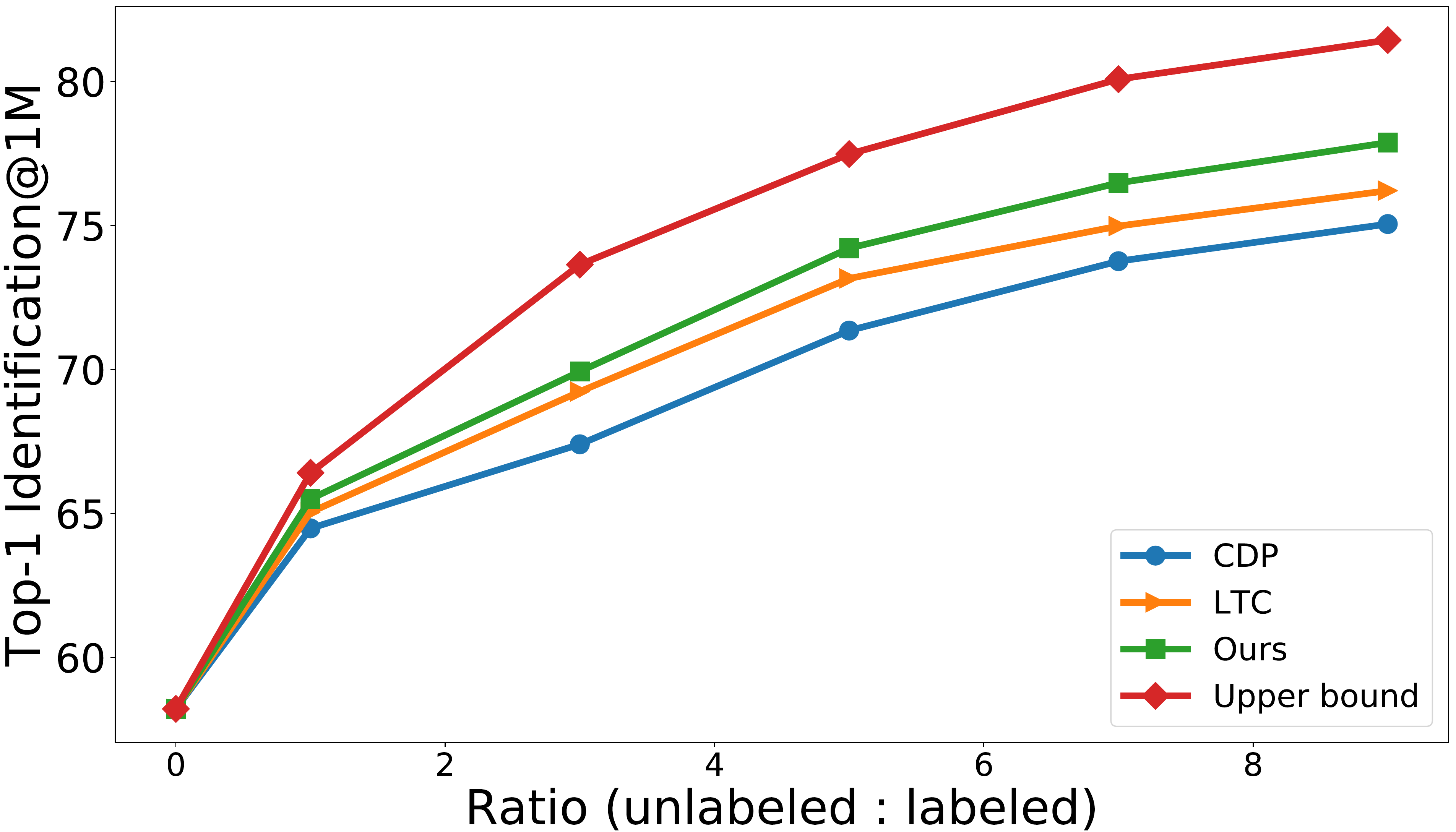}
	\caption{\small
        MegaFace top-1 Identification@1M.
    }
	\label{fig:exp_megaface}
\end{figure}


Following the pipeline of~\cite{zhan2018consensus,yang2019learning},
we apply the trained clustering model to assign pseudo labels to unlabeled
data, and leverage them to enhance the face recognition model.
As Sec.~\ref{sec:exp_setting} introduces, we split the dataset into 10
splits and randomly select 1 split to have the ground-truth labels, denoted as
$\cS_L$.
Particularly, the face recognition experiments involve 4 steps:
(1) use $\cS_L$ to train a face recognition model $\cM_r$;
(2) use $\cM_r$ to extract face features on $\cS_L$ and train the
clustering model $\cM_c$ with extracted features and corresponding labels
in $\cS_L$;
(3) use $\cM_c$ to assign pseudo labels to unlabeled images;
(4) use $\cS_L$ and unlabeled data with pseudo labels to train the final face
recognition model in a multi-task manner.
Note that $\cS_L$ is used to train both initial face
recognition model and face clustering model.

Different from previous work~\cite{zhan2018consensus,yang2019learning},
where the unlabeled data are assumed to be obtained sequentially and clustering
is performed 9 times on 9 splits separately,
we directly perform clustering on $5.21M$ unlabeled data,
which is more practical and challenging.
The upper bound is trained by assuming all unlabeled data have ground-truth
labels.
As Figure~\ref{fig:exp_megaface} indicates, all the three methods benefit from
an increase of the unlabeled data.
Owing to the performance gain in clustering, our approach outperforms
previous methods consistently and improves the performance of face recognition
model on MegaFace from $58.21$ to $77.88$.

\subsection{Ablation Study}

To study some important design choices,
we select MS-Celeb-1M($584K$) and DeepFashion for ablation study.


\vspace{-7pt}
\subsubsection{Confidence Estimator}
\label{sec:v_det}

\paragraph{Design of vertex confidence.}
We explore different designs of confidence.
As the confidence is related to the concept of ``density" described in Sec.
\ref{sec:related},
we first adopt two widely used unsupervised density as the
confidence~\cite{ester1996density,ankerst1999optics,rodriguez2014clustering}.
Given a radius, the first one is defined as the number of vertices and the
second one is computed by the sum of edge weights,
denoted them as $u_{num}^r$ and $u_{weight}^r$, respectively
as shown in Table~\ref{tab:exp_abl_v_det}.
Note that for these unsupervised definitions, the confidence is
directly computed without the learning process.
On the other hand, we can define various supervised confidence based
on the ground-truth labels. $s_{avg}$ is defined as the average
similarity to all vertices with the same label.
$s_{center}$ is defined as the similarity to the center, which is computed
as the average feature of all vertices with the same label.
$s_{nbr}$ is defined as Eq.~\ref{eq:v_importance}.
$s_{nbr}^F$ indicates using the top embedding $\mF_L$ to rebuild the graph.
To compare different confidence designs,
we adopt the same connectivity estimator by setting $\rho=0$ and $M=1$.
In this sense, the connectivity estimator directly selects the nearest
neighbors in the candidate set without learning.

As shown in Table~\ref{tab:exp_abl_v_det},
two unsupervised density definitions achieve relatively lower performance.
The assumption that high data density indicates high probability of
clusters may
not necessarily hold for all the situations. Besides, the performance is
sensitive to the selected radius for computing
density.
Table~\ref{tab:exp_abl_v_det} shows that the supervised confidence
outperforms the unsupervised confidence without the need to manually set a radius.
Among these three definitions, $s_{nbr}$ achieves better performance than $s_{avg}$ and
$s_{center}$.
As $s_{nbr}$ is defined on neighborhood, the learning of GCN may be easier compared
to $s_{avg}$ and $s_{center}$ which are defined with respect to all
samples in the same cluster.
In real practice, similar to saliency map fusion in saliency detection~\cite{goferman2011context,han2017cnns},
we can ensemble the outputs from different confidences to achieve better performance.

\begin{table}[t]
 \centering\small
 \caption{
	\small Design choice of vertex confidence.
	The confidence metrics are defined in Sec.~\ref{sec:v_det}.
	$\mF_L$ denotes the output feature embeddings of
$L$-th GCN layer in Sec.~\ref{sec:gcn_v}.
	}
 \begin{tabular}{l|c|cc|cc}
 \hline
 \multirow{2}{*}{Metric} &
 \multirow{2}{*}{$\mF_L$} &
 \multicolumn{2}{c|}{MS1M-584K} &
 \multicolumn{2}{c}{DeepFashion} \\ \cline{3-6}
 & & $F_P$ & $F_B$ & $F_P$ & $F_B$
\\\hline\hline
 $u_{num}^r$ & \texttimes & 61.65 & 64.8 & 19.42 & 45.85 \\\hline
 $u_{weight}^r$ & \texttimes & 81.78 & 80.47 & 29.31 & 52.81 \\\hline
 $s_{avg}$ & \texttimes & 82.37 & 83.32 & 30.11 & 56.62 \\\hline
 $s_{center}$ & \texttimes & 82.55 & 83.46 & 31.81 & 56.48 \\\hline
 $s_{nbr}$ & \texttimes & 82.76 & 83.61 & 32.24 & 57.11 \\\hline
 $s_{nbr}^F$ & \checkmark & 87.14 & 85.82 & 33.07 & 57.26\\\hline
 \end{tabular}
 \label{tab:exp_abl_v_det}
\end{table}



\vspace{-7pt}
\paragraph{Transformed embeddings.}
Comparison between $s_{nbr}$ and $s_{nbr}^F$ indicates that
using the transformed features to rebuild the affinity graph leads to
performance gain in both datasets.
This idea shares common concepts to Dynamic graph~\cite{wang2019dynamic}
where they rebuild the $K$NN graph after each graph convolutional layer.
However, on a massive graph with millions of vertices, constructing $K$NN graph per layer
will incur prohibitive computational budget.
The experiments indicate that only using the top embedding to rebuild the
graph can product reasonably well results.



\vspace{-9pt}
\subsubsection{Connectivity Estimator}
\label{sec:e_det}

\begin{figure}[t]
	\centering
	\includegraphics[width=0.7\linewidth]{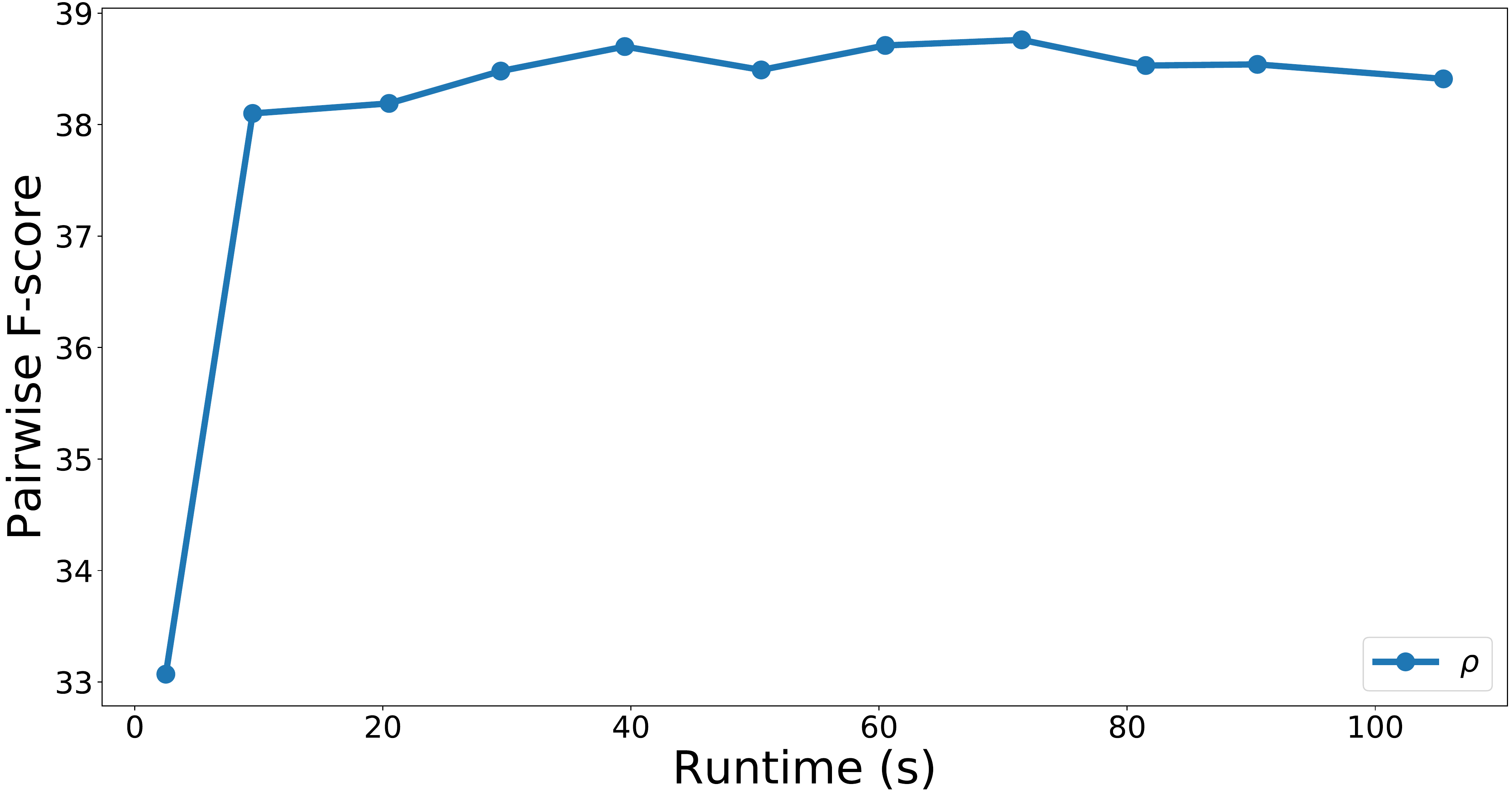}
	\caption{\small Influence of $\rho$ on DeepFashion.
    The leftmost point ($\rho=0$) indicates the result without GCN-E,
	while the rightmost point ($\rho=1$) employs GCN-E to all vertices.
    }
	\label{fig:exp_abl_rho}
\end{figure}

\vspace{-9pt}
\paragraph{The Influence of $\rho$.}
We vary $\rho$ from $0$ to $1$ with a step $0.1$.
As shown in Figure~\ref{fig:exp_abl_rho}, focusing only on
$10\%$ of vertices
with high confidence can lead to considerable performance gain while
adding very little computational cost.
As $\rho$ increases, more vertices benefit from the prediction of GCN-E
and thus $F_P$ increases.
There is a slight drop when applying GCN-E to all vertices, since
connections between unconfident vertices are often very complex,
and it may be hard to find common patterns for learning.
%

\vspace{-9pt}
\paragraph{The Influence of $M$.}
In Table below,
$M=-1$ indicates applying GCN-E without using the candidate set.
It includes unconfident neighbors, thus increasing the difficulty of learning
and leading to performance drop.
\begin{center}
{\small
 \begin{tabular}{c|c|c|c|c}
 \hline
 $M$ & -1 & 1 & 2 & 3
\\\hline
 $F_P$ & 29.85 & 38.47 & 1.19 & 0.31 \\
 $F_B$ & 56.12 & 60.06 & 56.43 & 52.46 \\\hline
\end{tabular}
}
\end{center}
When $M=1$, each vertex connects to its most connective neighbor in
the candidate set.
When $M > 1$, unconfident vertices will possibly connect to two
different clusters.
Although it increases the recall of obtained clusters, it may severely impair
the precision.
%

%



\section{Conclusion}
\label{sec:conclusion}

This paper has proposed a novel supervised face clustering framework, eliminating
the requirement of heuristic steps and numerous subgraphs.
The proposed method remarkably improves accuracy and efficiency on large-scale
face clustering benchmarks.
Besides, the experiments indicate the proposed approach generalizes well to
a testing set 10 times larger than the training set.
Experiments on fashion datasets demonstrate its potential application
to datasets beyond human faces.
In the future, an end-to-end learnable clustering framework is desired
to fully release the power of supervised clustering.

\vspace{7pt}
\noindent \textbf{Acknowledgement} This work is partially supported by the SenseTime Collaborative Grant on Large-scale Multi-modality Analysis (CUHK Agreement No. TS1610626 \& No. TS1712093),
the Early Career Scheme (ECS) of Hong Kong (No. 24204215), the General Research Fund (GRF) of Hong Kong (No. 14236516, No. 14203518 \& No. 14241716), and Singapore MOE AcRF Tier 1 (M4012082.020).

{\small
\bibliographystyle{ieee}
\bibliography{egbib}
}

\end{document}